\pdfoutput=1

\documentclass[11pt]{article}
\usepackage[utf8]{inputenc}
\usepackage{textgreek}

\usepackage[preprint]{acl}

\usepackage{times}
\usepackage{latexsym}
\usepackage{pgfplots}
\usepackage{booktabs}
\usepackage{multirow}
\usepackage{makecell}
\usepackage{array}
\usepackage[most]{tcolorbox}
\usepackage[T1]{fontenc}

\usepackage[utf8]{inputenc}
\usepackage{amsmath}
\usepackage{amssymb}
\usepackage{comment}

\usepackage{amsthm}

\newtheorem{definition}{Definition}
\usepackage{microtype}
\usepackage{float}
\usepackage{inconsolata}
\usepackage{multicol}
\usepackage{multirow}
\usepackage{graphicx}
\usepackage{xcolor}
\definecolor{JScolor}{RGB}{0, 100, 200}

\definecolor{DMcolor}{RGB}{200, 100, 0}

\definecolor{AOcolor}{RGB}{100, 0, 200}

\definecolor{BKcolor}{RGB}{75, 150, 75}

\definecolor{codegreen}{rgb}{0,0.0.2,0.6}
\definecolor{codegray}{rgb}{0.8,0.8,0.8}
\definecolor{codepurple}{rgb}{0.58,0,0.82}
\definecolor{backcolour}{rgb}{0.97,0.97,0.97}
\usepackage{listings}
\usepackage{xcolor}
\lstdefinestyle{mystyle}{
    backgroundcolor=\color{backcolour},   
    commentstyle=\color{codegreen},
    keywordstyle=\color{magenta},
    numberstyle=\tiny\color{codegray},
    stringstyle=\color{codepurple},
    basicstyle=\ttfamily\footnotesize,
    breakatwhitespace=false,         
    breaklines=true,                 
    captionpos=b,                    
    keepspaces=true,                 
    numbers=left,                    
    numbersep=5pt,                  
    showspaces=false,                
    showstringspaces=false,
    showtabs=false,                  
    tabsize=2
}
\title{(Almost) Free Modality Stitching of Foundation Models}

\author{
  \textbf{Jaisidh Singh\textsuperscript{1,2,4}},
  \textbf{Diganta Misra\textsuperscript{3,4}},
  \textbf{Boris Knyazev\textsuperscript{5}},
  \textbf{Antonio Orvieto\textsuperscript{3, 4, 6}}
\\
  \textsuperscript{1}University of Tübingen,
  \textsuperscript{2}Zuse School ELIZA
  \textsuperscript{3}ELLIS Institute Tübingen,
  \textsuperscript{4}MPI-IS Tübingen,
\\
  \textsuperscript{5}SAIT AI Lab Montréal,
  \textsuperscript{6}Tübingen AI Center
\\
\small{
  \textbf{Correspondence:} \href{mailto:jaisidh.singh@student.uni-tuebingen.de}{\tt{jaisidh.singh@student.uni-tuebingen.de}}
}
}

\begin{document}
\maketitle
\begin{abstract}
Foundation multi-modal models are often designed by stitching of multiple existing pretrained uni-modal models: for example, an image classifier with an text model. This \textit{stitching} process is performed by training a \emph{connector} module that aims to align the representation spaces of these uni-modal models towards a multi-modal objective. However, given the complexity of training such connectors on large scale web-based datasets coupled with the ever-increasing number of available pretrained uni-modal models, the task of uni-modal models selection and subsequent connector module training becomes computationally demanding. To address this under-studied critical problem, we propose \textbf{Hypernetwork Model Alignment (\textsc{Hyma})}, a novel all-in-one solution for optimal uni-modal model selection and connector training by leveraging hypernetworks. Specifically, our framework utilizes the parameter prediction capability of a hypernetwork to obtain jointly trained connector modules for $N \times M$ combinations of uni-modal models. In our experiments, \textsc{Hyma} reduces the cost of searching for the best performing uni-modal model pair by $10\times$, while matching the ranking and trained connector performance obtained via grid search across a suite of diverse multi-modal benchmarks.
\end{abstract}

\section{Introduction}\label{sec:intro}
Multi-modal foundation models have emerged as a new frontier in the Artificial Intelligence (AI) landscape. Fueled by the increasing need for considering inter-dependency of multiple data modalities in modern tasks, multi-modal foundation models often leverage modality-specific (uni-modal) models as sub-components, which are stitched together via a \emph{connector} module. A prominent class of such models is Vision-Language Models (VLMs)~\cite{clip,conclip,blip,flava}, which comprise image and text encoders that embed image and text concepts into a common contrastively learnt latent space. 
\begin{figure}[t]
    \centering
    \includegraphics[width=\linewidth]{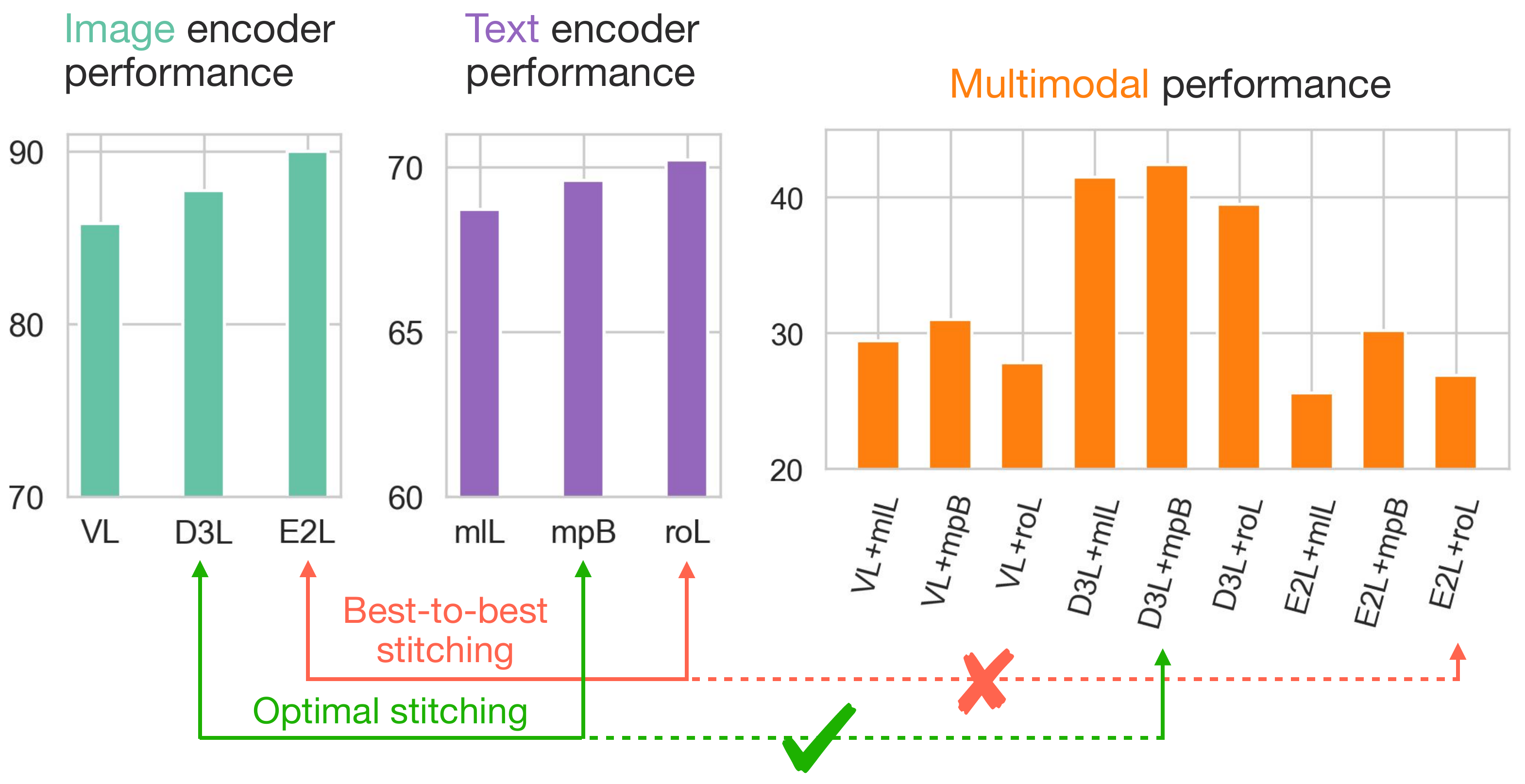}
    \caption{We train connectors between pretrained uni-modal models to show that uni-modal model performance is \textbf{not predictive} of multi-modal performance obtained by stitching. Image encoder performance refers to top-1 ImageNet-1K accuracy, text encoder performance refers to semantic search performance across 14 datasets~\cite{sbert}. Multi-modal scores refers to ImageNet-1K top-1 accuracy (classification by matching images to prompts such as ``\emph{this is a photo of a }\texttt{\{class\}})''.\protect\footnotemark}
    \vspace{-3mm}
    \label{fig:motiv}
\end{figure}
\footnotetext{All model abbreviations can be found in Appendix~\ref{sec:appendix}.}

\begin{figure*}[t]
    \centering
    \includegraphics[width=\linewidth]{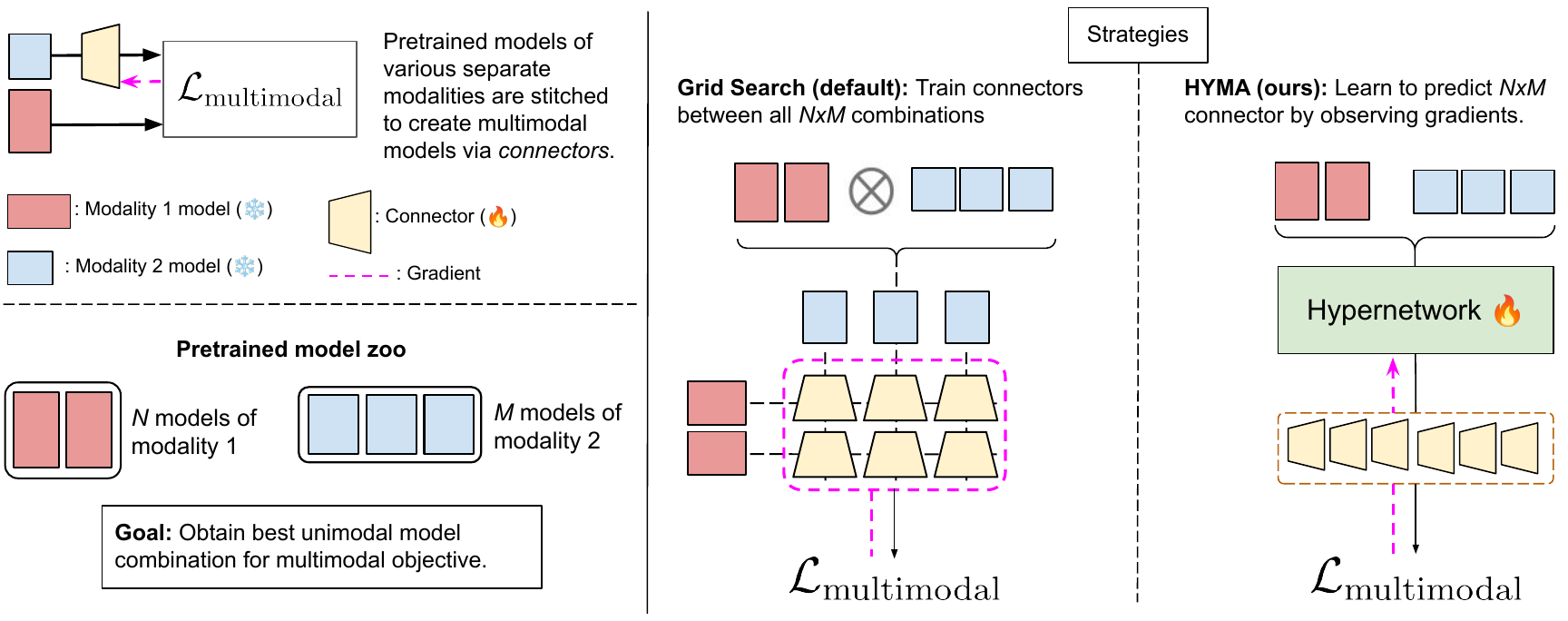}
    \caption{Given multiple options for uni-modal models, pair-wise grid search can be an expensive way to determine the best multi-modal combination. Alternatively, \textsc{Hyma} formulates search as a predictive or generative process.
    }
    \label{fig:overview*}
\end{figure*}

\emph{Connector} modules powering VLMs are often constructed as an $n$-layer multi-layer perceptron (MLP)~\cite{llava}, or in some cases even as simple as a linear layer~\cite{merullo2022linearly}, with the purpose of stitching modality-specific models. While some exceptions do arise where these modules are extensively engineered transformer-like architectures~\cite{blip2}, the vast majority consensus on the design of such connector modules has been limited to MLPs~\cite{zhu2025connector} due to their efficiency. 

While training connector modules for a pair of predetermined uni-modal models is feasible, the picture becomes more complex when considering multiple uni-modal options and aiming to optimize for downstream performance after stitching. Indeed, it is often not the case~(see Figure~\ref{fig:motiv}) that simply choosing to align best-performing uni-modal models leads to the best multi-modal performance. This trend is further illustrated in Table~\ref{tab:size_motivation}, where uni-modal model parametric capacity fails to serve as a reliable predictor of multi-modal performance. Consequently, the cost of optimal stitching can grow quadratically with the number of available options on both ends. In addition, the availability of extremely large web-scale pretraining datasets, consisting of samples in the order of billions~\cite{laion5b,cc12m,desai2021redcaps}, constitutes a blocker for proper ablation on such design choices. 

\begin{table}[H]
\centering
    \resizebox{\linewidth}{!}{
    \begin{tabular}{@{}l l c c@{}}
    \toprule
    \textbf{\textsc{\textbf{I}} (\#Params)} & \textbf{\textsc{\textbf{T}} (\#Params)}  & \textbf{Total \#Params} & \textbf{Perf.} \\
    \midrule
    EVA2-L (305M) & roberta-L (355M) & \textbf{660M $+\, c$} & 26.85 \\
    DeiT3-L (304M) & mpnet-B (109M) & 413M $+\, c$ & \textbf{42.63} \\
    \bottomrule
\end{tabular}}
\caption{\textbf{Parametric capacity of unimodal models is not a reliable indicator of multimodal performance}. On the task of multi-modal image classification using the ImageNet-1k dataset, we observe that stitching the highest-capacity models: EVA-2 Large (305M) for the image modality (\textbf{\textsc{I}}) and RoBERTa Large (355M) for the text modality (\textbf{\textsc{T}}), totaling 660M + $c$ parameters—yields significantly lower performance than a smaller stitched pair: DeiT-3 Large (\textbf{\textsc{I}}) (304M) and MPNet-Base (\textbf{\textsc{T}}) (109M), totaling just 413M + $c$ parameters. $c$ denotes the parameters contributed by 1-hidden layer MLP connector and \textbf{Perf.} denotes the Top-1 accuracy metric.}
\label{tab:size_motivation}
\end{table}

We highlight and define the problem, which we term \textbf{Multi-modal Optimal Pairing and Stitching (M-OPS)}, as:
\begin{itemize}
    \item \textbf{Pairing}: Given a set of $N$ models in modality 1 (e.g., vision) and $M$ models in modality 2 (e.g., text), provide the optimal (best performing) combination pair ($n,m; \ n \in N \ | \ m \in M$) to construct a multi-modal model for a target task and/or under target constraints (e.g., parametric size, embedding dimensions).
    \item \textbf{Stitching}: For the selected uni-modal models ($n,m$), obtain the optimal trained connector $f_\theta$ that stitches them to construct the target multi-modal model.
\end{itemize}

Due to the infeasibility of addressing the \textbf{pairing} sub-problem of \textbf{M-OPS} via a grid-search approach for a large $N \times M$ pair, we propose a novel alternative approach to tackle both the pairing and stitching steps in a single unified manner that utilizes a HyperNetwork~\cite{hypernetworks}. The key idea behind our approach is that stitching similar models shares latent semantics, which can be captured by jointly training a network to generate connectors.

We present \textbf{Hypernetwork Model Alignment (\textsc{Hyma})}, a method that, given $N$ modality 1~(e.g., image) and $M$ modality 2~(e.g., text) models, leverages a hypernetwork~\cite{hypernetworks} that jointly learns to generate connectors for all possible $N\times M$ combinations. Our approach serves both as an indicator for optimal model pair configurations and as a trainer that produces \emph{stitched} multi-modal models performing on par with the best stitched model pair obtained via grid search. In our experiments, where $N \times M$ can be as high as 27 (discussed in Section~\ref{sec:vlm_exp}), our method enables an efficiency gain of $10\times$ in obtaining the best stitched model pair compared to grid search. 

We highlight our contributions as follows:
\begin{enumerate}
    \item We propose \textbf{Hypernetwork Model Alignment (\textsc{Hyma})}, a hypernetwork-based approach for obtaining strong uni-modal model pairs that perform on par with the best stitched model pair obtained via grid search at an order-of-magnitude lower computational cost.
    \item Our proposed approach \textbf{\textsc{Hyma}} is, to the best of our knowledge, the first to demonstrate the effectiveness of hypernetworks for solving the \textbf{M-OPS} problem defined above.
    \item We empirically demonstrate the performance and efficiency of \textbf{\textsc{Hyma}} on VLMs across various multi-modal benchmarks.
\end{enumerate}

\section{Background}\label{sec:bg}

In this section, we present the necessary preliminaries for the \textbf{M-OPS} problem, along with the general training paradigm of hypernetworks. These formal definitions establish the foundation for our proposed method, \textsc{Hyma}, which we introduce in the following section.

\begin{definition}[Hypernetworks for Parameter Prediction]
A hypernetwork~\cite{hypernetworks} is a neural network $H_\phi$ parameterized by $\phi$, designed to predict the parameters $\theta$ of a target network $f_\theta$ based on a conditioning input $\mathbf{c}$. The parameter generation process is defined as:
\[
H_\phi(\mathbf{c}) = \theta.
\]
The parameters $\phi$ of the hypernetwork are optimized indirectly via the performance of the generated network $f_\theta$ on a downstream task. Given a task-specific loss $\mathcal{L}_{\text{task}}$ evaluated on corresponding data, the optimization objective becomes:
\[
\phi^* = \arg\min_{\phi} \mathcal{L}_{\text{task}}(f_{H_\phi(\mathbf{c})}).
\]
The trained hypernetwork $H_{\phi^*}$ can then be used to generate task-adapted parameters $\theta$ for $f$ given new conditioning inputs. Optimizing $\phi$ rather than $\theta$ directly can offer advantages in terms of training dynamics, capacity control, and generalization~\cite{chauhan2024brief}.
\end{definition}
\begin{figure*}[t]
    \centering
    \includegraphics[width=0.95\linewidth]{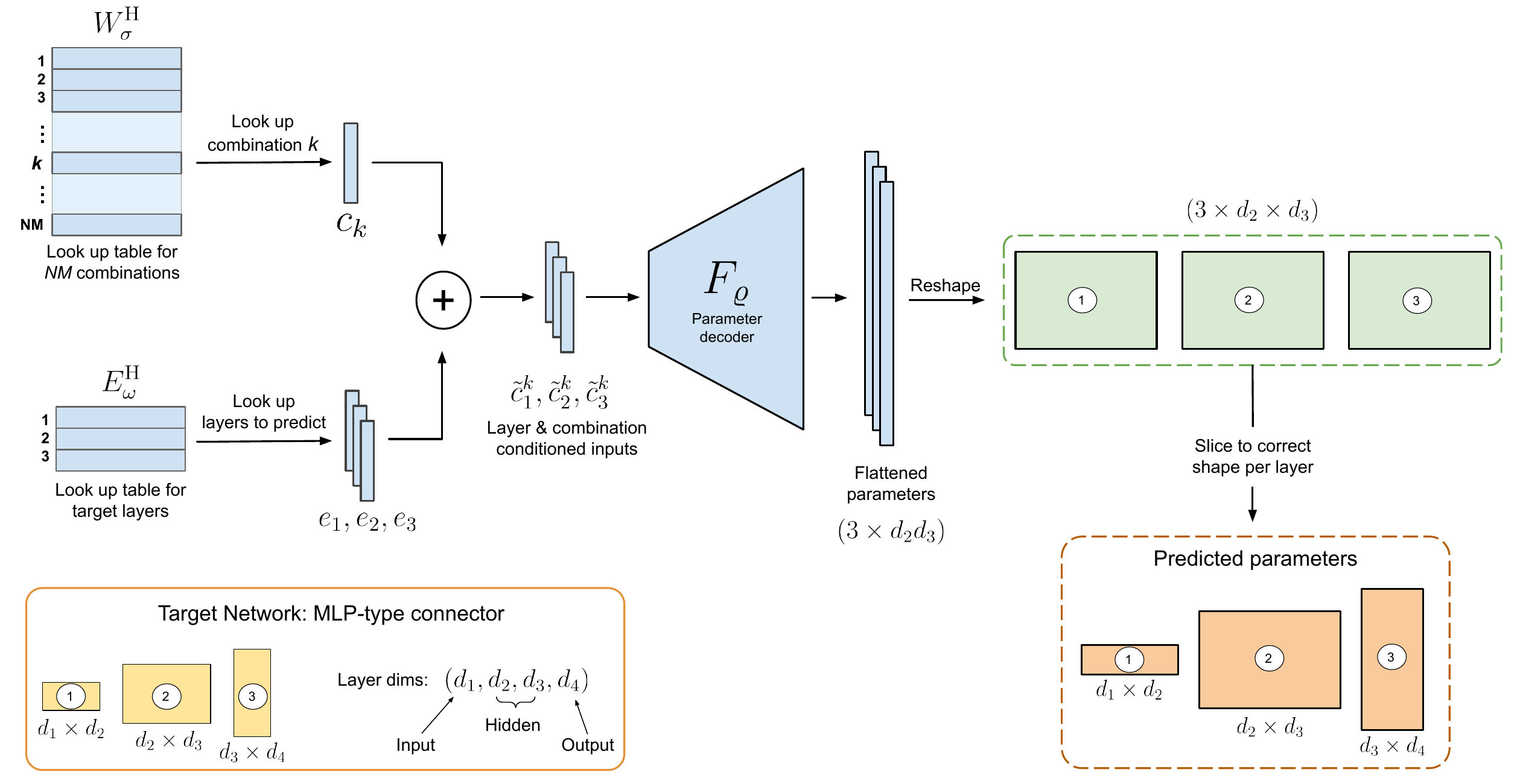}
    \caption{A visual walkthrough of our hypernetwork architecture is provided above. We take the example of predicting the parameters of an MLP-type connector with depth$=3$ (denotes $2$ hidden layers).}
    \label{fig:hyma_arch}
\end{figure*}

For simplicity, assume encoders producing sequences of $P$ features~(e.g., number of patches or tokens) living in a $ D$-dimensional space.
\begin{definition}[Connector-based multi-modal stitching]
Let $\mathcal{A} : \mathcal{X}_A \rightarrow \mathbb{R}^{  D_A}$ and $\mathcal{B} : \mathcal{X}_B \rightarrow \mathbb{R}^{ D_B}$ be pretrained uni-modal encoders for two different modalities with input spaces $\mathcal{X}_A$ and $\mathcal{X}_B$, respectively. The goal is to construct a multi-modal model by learning a connector function $f_\theta : \mathbb{R}^{ D_A} \rightarrow \mathbb{R}^{ D_B}$ that stitches the output of $\mathcal{A}$ to the representation space of $\mathcal{B}$: given input pairs $(\mathbf{u}, \mathbf{v}) \in \mathcal{X}_A \times \mathcal{X}_B$, the connector stitches the modality-A features $$\mathbf{x}^a = \mathcal{A}(\mathbf{u}) \in \mathbb{R}^{ D_A}$$ to modality-B space via
\[
\tilde{\mathbf{x}}^a = f_\theta(\mathbf{x}^a) \in \mathbb{R}^{ D_B}
\]

The stitched representation $\tilde{\mathbf{x}}^a$ is then combined with $\mathbf{x}^b = \mathcal{B}(\mathbf{v})$ to construct a joint multi-modal representation. The connector parameters $\theta$ are optimized while keeping $\mathcal{A}$ and $\mathcal{B}$ frozen. The training objective follows  \textbf{contrastive stitching}, that uses a similarity function $\text{sim}(\cdot, \cdot)$ and temperature $\tau$ to train the connector on the InfoNCE~\cite{infonce} loss~(quadratic):
    \[
    \mathcal{L}_{\text{contrastive}}(\theta) = -\log \frac{\exp(\text{sim}(\tilde{\mathbf{x}}^a, \mathbf{x}^b)/\tau)}{\sum_{j} \exp(\text{sim}(\tilde{\mathbf{x}}^a, \mathbf{x}^b_j)/\tau)}
    \]
\end{definition}

\section{Methodology}\label{sec:method}

\subsection{Problem formulation}
We aim to jointly learn $N \times M$ connectors, where each connector is specified to the hypernetwork via a conditional input $\mathbf{c}^k$. More formally, for the $k^{th}$ model combination, the hypernetwork generates the parameters as $H_\phi(\mathbf{c}^k)$. The resulting connector $f_{H_{\phi}(\mathbf{c}^k)}$ is then used to compute a task-specific loss. The overall training loss is computed by averaging over all combinations:
\begin{equation}
    \mathcal{L}_{\text{\textsc{Hyma}}} = \frac{1}{NM} \sum_{k=1}^{NM} \mathcal{L}_{\text{task}}(f_{H_{\phi}(\mathbf{c}^k)}).
\end{equation}
Here, $\mathcal{L}_{\text{task}}$ corresponds to a contrastive InfoNCE loss (for retrieval-style objectives like that in CLIP~\cite{clip}). The trained hypernetwork is denoted by $H_{\phi^*}$, where $\phi^* = \arg\min_{\phi} \mathcal{L}_{\text{\textsc{Hyma}}}$. Following prior work~\cite{ape,tinyllava}, we restrict connectors to be multi-layer perceptrons (MLPs).

\subsection{Hypernetwork architecture}

We define the hypernetwork as a function $H_\phi: \mathbb{R}^{C} \to \mathbb{R}^{D_\theta}$, mapping conditional inputs $\mathbf{c} \in \mathbb{R}^C$ to connector parameters $\theta \in \mathbb{R}^{D_\theta}$. We describe next how $\mathbf{c}$ is constructed and how it is mapped to the parameter space.

\paragraph{Conditional inputs:} We use a learnable lookup table of embeddings $\mathbf{W}^{\text{H}}_\sigma \in \mathbb{R}^{NM \times C}$, where $\mathbf{c}^k = \mathbf{W}^{\text{H}}_\sigma[k]$ encodes the $k^{th}$ model pair.
%
%
%

\paragraph{Mapping conditional inputs to parameters:} The hypernetwork $H_\phi$ is implemented using an MLP $F_\varrho$, which predicts connector parameters layer-wise. Each layer prediction is conditioned on both $\mathbf{c}^k$ and a learnable layer-specific embedding $\mathbf{e}_j = \mathbf{E}^{\text{H}}_\omega[j]$, such that:
\[
F_\varrho(\tilde{\mathbf{c}}^k_j) \in \mathbb{R}^{D_{\vartheta^k}}, \quad \text{where} \quad \tilde{\mathbf{c}}^k_j = \mathbf{c}^k + \mathbf{e}_j
\]
and $\vartheta^k$ denotes the size of the largest layer in the $k^{th}$ connector. The output is then sliced to the appropriate dimension for layer $j$. This process is repeated for all layers, and the resulting parameters are concatenated to form the complete connector parameter vector $\theta^k \in \mathbb{R}^{D_\theta^k}$. This modular, layer-wise parameterization makes the hypernetwork more tractable and memory-efficient.

\subsection{Mini-batching model combinations for scalable hypernetwork training}

Jointly training connectors for all $N \times M$ model combinations can become computationally prohibitive. To address this, we follow the strategy of model mini-batching~\cite{ghn3}, wherein each training step operates over a batch of $B_m$ model combinations. The modified loss is:\\

\begin{equation}
\mathcal{L}_{\text{\textsc{Hyma}}} = \frac{1}{B_m} \sum_{k=1}^{B_m} \mathcal{L}_{\text{task}}(f_{H_{\phi}(\mathbf{c}^k)}).
\label{eq:model_batch}
\end{equation}
Each training step proceeds as follows:
\begin{enumerate}
    \item Sample a data batch of size $B_d$.
    \item For each data sample, evaluate $\mathcal{L}_{\text{\textsc{Hyma}}}$ over each of the $B_m$ model combinations.
    \item Use the accumulated loss to update hypernetwork parameters $\phi$.
\end{enumerate}
This training strategy enables \textsc{Hyma} to scale efficiently without requiring all models or their combinations to be loaded simultaneously.
\begin{figure*}[t]
    \centering
    \includegraphics[width=0.49\linewidth]{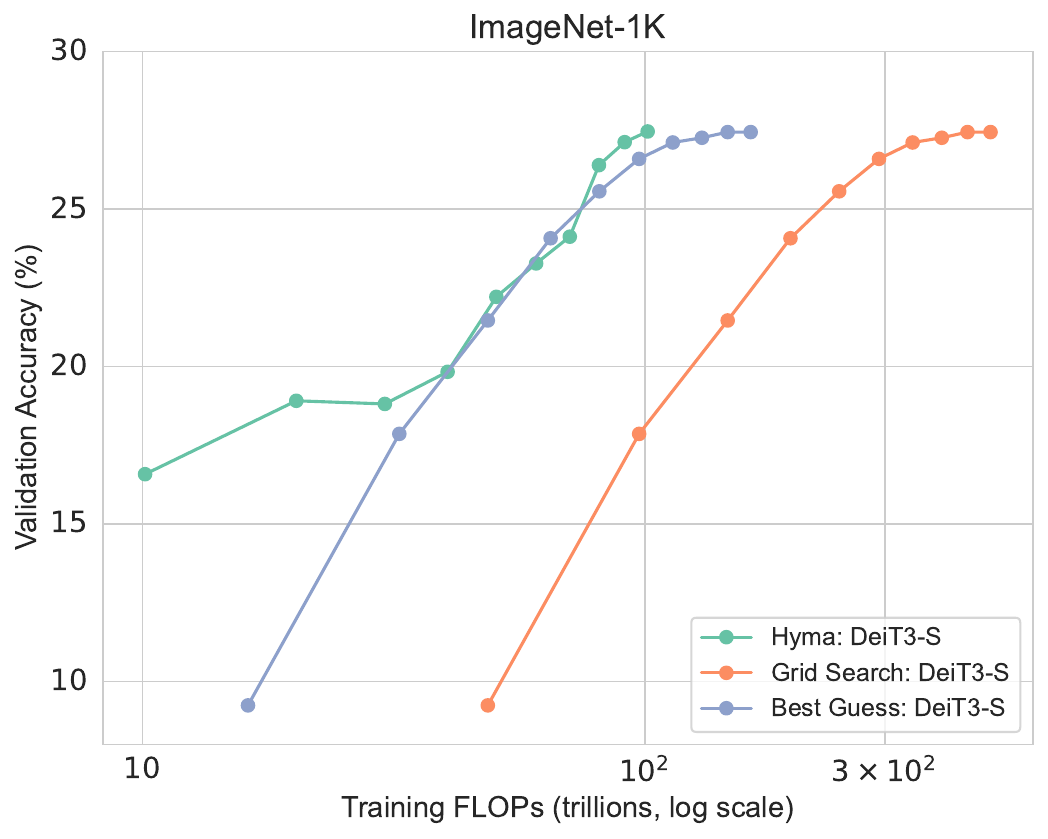}
    \includegraphics[width=0.49\linewidth]{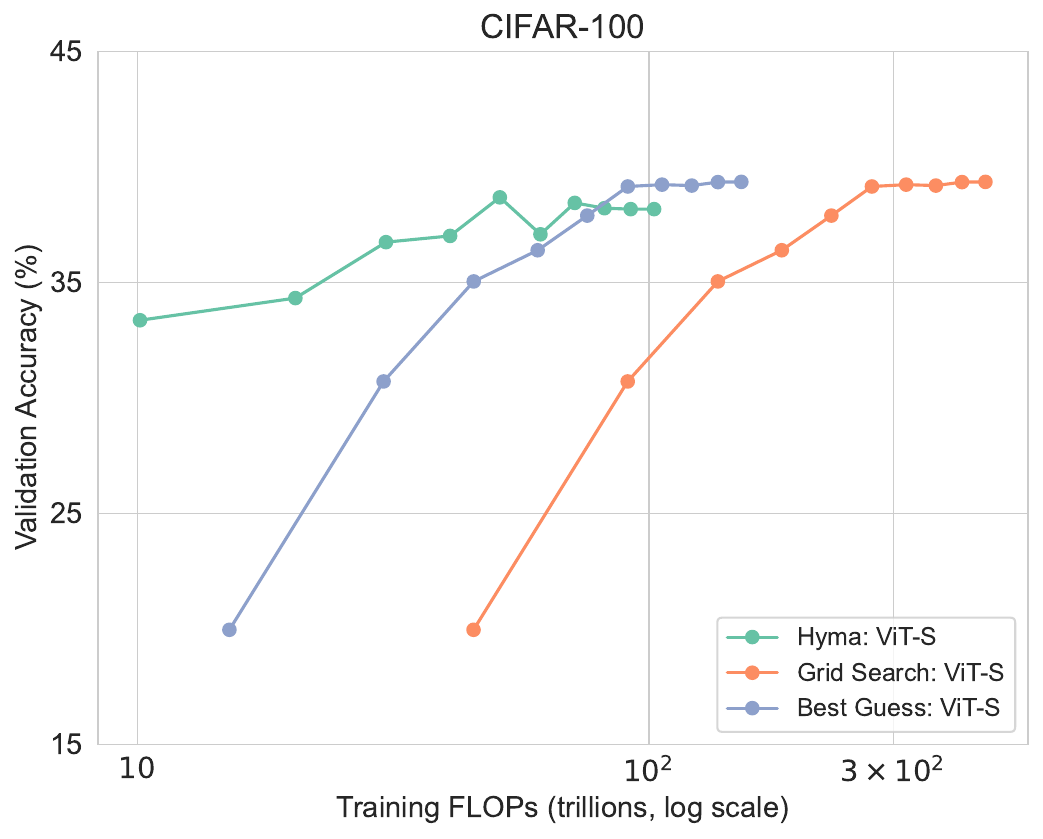}
    \caption{\textbf{MLP${_1}$ | $N \times M =3$}: We show the trade-off between computational resources (measured in FLOPs) and performance of the best stitched model pairs across all comparative baselines. We find that \textsc{Hyma} is able to predict a highly performance pairing at a significantly reduced FLOP cost in comparison to training on the optimal model pair as well as search over all model pairs for $N\times M=3$.}
    \label{fig:vlm_e1}
\end{figure*}
\section{Experiments}
\subsection{Baselines} \label{sec:baselines}
To ensure comprehensive evaluation of our proposed method, we compare against the following baselines:

\begin{itemize}
    \item \textbf{Random}: A naive baseline that randomly selects and stitches uni-modal model pairs using the specified connector on the target multi-modal dataset. Reported performance is the average over \textit{five} independent trials.
    
    \item \textbf{UniModal Top-1 (UniT-1)}: Inspired by the observation in Fig.~\ref{fig:motiv}, this baseline stitches the top-performing individual uni-modal models—selected based on their uni-modal benchmark performance—via the target connector. For VLMs, image models are ranked by ImageNet Top-1 accuracy, and text models by their corresponding sentence embedding performance. 
    
    \item \textbf{Ask-LLM}: Since uni-modal model properties such as parameter count and pretraining data can influence multi-modal performance, we define a baseline \texttt{Ask-LLM}. Here, a language model is prompted with metadata from the model zoo for both modalities and asked to select the most suitable pair for the target task. The chosen pair is stitched using a connector and evaluated in isolation.
    
    \item \textbf{AutoPair}: To enable a fair comparison with \textsc{Hyma}'s efficiency-focused design, we implement an pairing baseline that iteratively searches a given set of pairs by training for a fixed number of epochs, and then prunes all pairs below the median performance. Specifically, AutoPair optimizes model pair selection and stitching within a FLOPs budget equal to that used by \textsc{Hyma} for the same model zoo. More details are provided in Section~\ref{app:AP}. 
    
    \item \textbf{\textcolor{blue}{Oracle (Grid Search)}}: This upper-bound baseline performs exhaustive grid search over all model pairs in the zoo, independently training and evaluating each stitched pair. While this provides optimal performance, it is computationally prohibitive.

    \item \textbf{Best Guess}: A hypothetical upper-bound baseline representing the training cost of the model combination that would yield the best multi-modal pair after stitching, assuming the optimal pair was known in advance.
\end{itemize}

\subsection{Models}
All model details are provided in Appendix~\ref{sec:appendix}. To construct our Vision-Language Models (VLMs), we define a model zoo containing $N = 9$ image encoders: \texttt{ViT-S, DeiT-S, DeiT3-S, ViT-B, DeiT-B, DeiT3-B, ViT-L, DeiT3-L, Eva2-L} and $M = 3$ text encoders: \texttt{minilm-L, mpnet-B, roberta-L}. This results in a total of $N \times M = 27$ possible VLM configurations.

\subsection{Connector variants}
We test \textsc{Hyma} against the aforementioned baselines across three connector configurations:
\begin{enumerate}
    \item \textbf{Linear}: As demonstrated in \cite{merullo2022linearly}, we construct the connector to be a linear layer parameterized via $\theta$, mapping from the embedding space of the text encoder to the image encoder of a specific pair.
    \item \textbf{MLP{$_1$}}: An MLP with one hidden layer of hidden dimension set $1024$. 
    \item \textbf{MLP{$_2$}}: A MLP with two hidden layers, each of dimension $1024$.
\end{enumerate}


\subsection{Datasets}

We employ the LLaVA-CC558K dataset~\cite{tinyllava}, which consists of $558{,}128$ high-quality synthetic image-text pairs. Connectors between image and text encoders are trained using the contrastive InfoNCE loss~\cite{infonce} for $10$ epochs, after which the best-performing checkpoint is selected. Hyperparameters are tuned for performance, stability, and GPU efficiency, detailed in Appendix. 


\subsection{Evaluation Tasks}
Post-training, the resulting VLMs are evaluated on the following four downstream tasks:
\begin{itemize}
    \item \textbf{Multi-modal Image Classification (MIC)}: We compute the zero-shot top-1 image classification accuracies of the VLMs on the ImageNet-1K~\cite{imagenet} and the CIFAR-100~\cite{cifar100} datasets. The evaluation follows an image-text matching approach, where the text corresponding to each image input takes the form: \textsc{“This is a photo of a \{class\}”}.
    \item \textbf{Image-Text Matching (ITM)}: Here, we compute the zero-shot recall$\ @ \ 5$ scores of the VLMs on the MSCOCO validation split~\cite{mscoco} and the Flickr-8K~\cite{flickr8k} datasets.
    \item \textbf{Visual Question Answering (VQA)}: We use the validation splits of the OK-VQA~\cite{okvqa} and the Text-VQA~\cite{textvqa} datasets. Implementation details for VQA are given in Appendix~\ref{app:vqa}.
\end{itemize}

\section{Empirical Results}\label{sec:vlm_exp}
\subsection{MLP{$_1$} | $\mathbf{N \times M=3}$}
\label{sec:vlm_exp1_s}
Initially, we stitch  $N=3$ image encoders (\texttt{ViT-S}, \texttt{DeiT-S}, \texttt{DeiT3-S}) with $M=1$ text encoder (\texttt{MiniLM}) using an MLP connector of 1 hidden layer (MLP$_1$). This yields a total of $N\times M=3$ possible VLMs, that we construct and evaluate on the image-classification task. For the best performing combination per evaluation benchmark, we show its accuracy in Figure~\ref{fig:vlm_e1} as well as the computational resources, measured in floating point operations (FLOPs) required to obtain the corresponding connectors.

On ImageNet-1K, \texttt{DeiT3-S} emerges as the best image encoder to be stitched with \texttt{minilm-L}. Further, \textsc{Hyma} and \emph{Grid Search} (and \emph{Best Guess}) exhibit the same final performance, i.e., $27.4$ \% top-1 accuracy. On the other hand, the most performative image encoder when stitched to \texttt{MiniLM} is \texttt{ViT-S}. In terms of performance, \textsc{Hyma} exhibits a top-1 accuracy of $38.4$ \%, nearly matching the performance of baselines that individually train connectors to find the optimal setting, i.e., $39.3$ \%.
Also, \textsc{Hyma} is strongly cost-effective for VLMs, being $4.44\times$ and $1.48\times$ more compute-efficient than \emph{Grid Search} and \emph{Best Guess} respectively.
\begin{table}[H]
    \centering
    \small
    \resizebox{\linewidth}{!}{
    \begin{tabular}{l|lr|lr}
    \toprule
        \multirow{2}{*}{\textbf{Dataset}} & \multicolumn{2}{c|}{\textbf{Efficiency $\mathbf{@10}$ ep ($\times$)}} & \multicolumn{2}{c}{\textbf{Efficiency $\mathbf{@}$ best ($\times$)}} \\
      & \textbf{BG} & \textbf{GS} & \textbf{BG} & \textbf{GS}\\
     \midrule
     IN-1K & 1.48 & 4.44 & 1.48 & 4.44\\
     CIFAR-100 & 1.48 & 4.44 & 2.96 & 8.89\\
     \bottomrule
    \end{tabular}}
    \caption{$N\times M = 3$, MLP$_1$: \textsc{Hyma} is significantly more compute-efficient than independently stitching model pairs, as shown w.r.t \textit{Best Guess (BG)} and \textit{Grid Search (GS)}.}
    \label{tab:poc_eff_tab}
\end{table}
\begin{table*}
    \centering
    \small
    \resizebox{\textwidth}{!}{%
    \begin{tabular}{l|c|c|ccc|c|cccc}
    \toprule
       \multirow{2}{*}{\textbf{Task}} & \multirow{2}{*}{\textbf{Dataset}} & \multirow{2}{*}{\textbf{Connector}} & \multicolumn{3}{c|}{\textbf{NDCG $\boldsymbol{@\ k} \ (\uparrow)$}} & \textbf{$\boldsymbol{\rho} \ (\uparrow)$} & \multicolumn{4}{|c}{\textbf{$\Delta_{\texttt{Performance}}\ (\uparrow)$}} \\
       & & & ${k=5}$ & ${k=7}$ & ${k=10}$ & {$N\times M=27$} & Random & UniT-1 & Ask-LLM & \textcolor{blue}{Oracle (GS)} \\
        \midrule
        \multirow{6}{*}{\textbf{MIC}} & \multirow{3}{*}{IN-1K} & Linear   & 1.0 & 1.0 & 0.98 & 0.97 & \textcolor{teal}{+}6.93 & \textcolor{teal}{+}13.51 & \textcolor{teal}{+}13.51 & \textcolor{magenta}{-}4.14\\
        & &  MLP{$_1$} & 1.0 & 0.98 & 0.96 & 0.91 & \textcolor{teal}{+}4.78 & \textcolor{teal}{+}11.11 & \textcolor{teal}{+}11.11 & \textcolor{magenta}{-}4.47 \\
        & &  MLP{$_2$} & 0.96 & 0.93 & 0.92 & 0.89 &\textcolor{teal}{+}3.89 & \textcolor{teal}{+}10.34 & \textcolor{teal}{+}10.34 & \textcolor{magenta}{-}5.91 \\
        \cmidrule{2-11}
        & \multirow{3}{*}{CIFAR-100} & Linear   & 0.88 & 0.96 & 0.97 & 0.97 &\textcolor{teal}{+}6.91 & \textcolor{teal}{+}38.50 & \textcolor{teal}{+}38.50 & \textcolor{magenta}{-}3.73 \\
        & &  MLP{$_1$} & 0.83 & 0.96 & 0.97 & 0.86 & \textcolor{teal}{+}6.31 & \textcolor{teal}{+}35.21 & \textcolor{teal}{+}35.21 & \textcolor{magenta}{-}1.85 \\
        & &  MLP{$_2$} & 0.74 & 0.93 & 0.95 & 0.90 & \textcolor{teal}{+}5.01 & \textcolor{teal}{+}35.48 & \textcolor{teal}{+}35.48 & \textcolor{magenta}{-}3.06 \\
        \midrule
        \multirow{6}{*}{\textbf{ITM}} & \multirow{3}{*}{MSCOCO} & Linear   & 0.96 & 0.95 & 0.99 & 0.99 & \textcolor{teal}{+}4.94 & \textcolor{teal}{+}31.62 & \textcolor{teal}{+}33.20 & \textcolor{magenta}{-}2.0 \\
        & &  MLP{$_1$} & 0.92 & 0.91 & 0.97 & 0.99 & \textcolor{teal}{+}3.72 & \textcolor{teal}{+}28.41 & \textcolor{teal}{+}28.41 & \textcolor{magenta}{-}3.06 \\
        & &  MLP{$_2$} & 0.96 & 0.91 & 0.97 & 0.98 & \textcolor{teal}{+}2.22 & \textcolor{teal}{+}27.30 & \textcolor{teal}{+}27.30 & \textcolor{magenta}{-}4.03 \\
        \cmidrule{2-11}
        & \multirow{3}{*}{Flickr-8K} & Linear   & 0.95 & 0.99 & 0.99 & 0.99 & \textcolor{teal}{+}5.18 & \textcolor{teal}{+}26.68 & \textcolor{teal}{+}7.83 & \textcolor{magenta}{-}2.06 \\
        & &  MLP{$_1$} & 1.0 & 1.0 & 0.99 & 0.99 & \textcolor{teal}{+}3.54 & \textcolor{teal}{+}23.32 & \textcolor{teal}{+}23.32 & \textcolor{magenta}{-}2.26 \\
        & &  MLP{$_2$} & 0.92 & 0.99 & 0.96 & 0.98 & \textcolor{teal}{+}1.92 & \textcolor{teal}{+}21.44 & \textcolor{teal}{+}21.44 & \textcolor{magenta}{-}3.25 \\
        \midrule
        \multirow{6}{*}{\textbf{VQA}} & \multirow{3}{*}{OK-VQA} & Linear & 0.95 & 0.95 & 0.98 & 0.99 & \textcolor{teal}{+}0.81 & \textcolor{teal}{+}7.86 & \textcolor{teal}{+}7.86 & \textcolor{magenta}{-}0.43\\
        & & MLP${_1}$ & 0.94 & 0.90 & 0.95 & 0.95 & \textcolor{teal}{+}0.49 & \textcolor{teal}{+}6.63 & \textcolor{teal}{+}6.63 & \textcolor{magenta}{-}0.77 \\
        & & MLP${_2}$ & 0.99 & 0.93 & 0.93 & 0.97 & \textcolor{teal}{+}0.01 & \textcolor{teal}{+}6.81 & \textcolor{teal}{+}6.81 & \textcolor{magenta}{-}1.44 \\
        \cmidrule{2-11}
        & \multirow{3}{*}{Text-VQA} & Linear & 0.94 & 0.97 & 0.99 & 0.97 & \textcolor{teal}{+}1.31 & \textcolor{teal}{+}3.64 & \textcolor{teal}{+}3.64 & \textcolor{magenta}{-}0.06 \\
        & & MLP${_1}$ & 0.92 & 0.87 & 0.90 & 0.87 & \textcolor{teal}{+}0.72 & \textcolor{teal}{+}2.59 & \textcolor{teal}{+}2.59 & \textcolor{magenta}{-}0.32 \\
        & & MLP${_2}$ & 0.85 & 0.87 & 0.86 & 0.87 & \textcolor{teal}{+}0.72 & \textcolor{teal}{+}2.28 & \textcolor{teal}{+}2.28 & \textcolor{magenta}{-}0.59 \\
        \bottomrule
    \end{tabular}%
    }
    \caption{\textbf{{\textsc{Hyma}} VLM Results}: We report the ranking similarity between {\textsc{Hyma}} and the Oracle—Grid Search (GS)—using NDCG and Spearman’s $\rho$. Across all three connector configurations, {\textsc{Hyma}} exhibits a strong correlation with GS rankings. Additionally, we show the performance gain ($\Delta$) of the best connector obtained post stitching via {\textsc{Hyma}}, compared to four baselines: (a) \textit{Random}: Random pairing and stitching (averaged over five runs), (b) \textit{UniT-1}: Stitching the best unimodal models based on unimodal benchmarks, (c) \textit{Ask-LLM}: Stitching based on model pairs selected via prompting Claude 4 Sonnet (detailed prompt in appendix), and (d) \textit{Oracle}: Full grid search over all possible configurations on the complete model zoo ($N \times M = 27$).}
    \label{tab:vlm_tab}
\end{table*}

\subsection{Linear, MLP{$_1$}, MLP{$_2$} | $\mathbf{N \times M=27}$}
\label{sec:vlm_e2}
After demonstrating the efficacy of \textsc{Hyma} on a small search space of $N\times M=3$ combinations and for MLP$_1$ scale up the number of combinations in comparison to $N\times M=27$, and vary the capacity of the connectors in use (Linear, MLP$_1$, MLP$_2$). This yields $81$ total VLMs. Table~\ref{tab:vlm_tab} shows the performance of \textsc{Hyma} in terms of a search, i.e., how well it matches the true ranking given by full grid search. Performance gain ($\Delta$) is also reported across the \textit{Random, UniT-1, Ask-LLM}, and \textcolor{blue}{\textit{Oracle (GS)}} baselines for each task and dataset employed.
\paragraph{Multi-modal Image classification:} For multi-modal image classification on the ImageNet-1K, we find that the ranking order of the stitching performed by \textsc{Hyma} reflects that found by full grid search to strong extent. This is indicated by the normalized discounted cumulative gain (NDCG $@ \ k$) computed for the top 5 and 7 ranks. Additionally, Spearman's $\rho$ across all $N\times M=27$ ranks further corroborates this. Notably, both NDCG $@ \ k$ and Spearman's $\rho$ for CIFAR-100 are lower in value w.r.t ImageNet-1K. In terms of performance gains, \textsc{Hyma} improves upon random selection of encoder pairs to stitch, as well as selecting encoders based on their uni-modal performance. Interestingly, we find that asking a massively pretrained LLM such as Claude 4 Sonnet yields a similar result to \textit{UniT-1}. For \textcolor{blue}{Oracle (GS)}, find that the best stitchings generated by \textsc{Hyma} underperform average of $4.84 \ \%$ and $2.88$ for ImageNet-1K and CIFAR-100 across all connector types. However, this occurs at $10 \times$ fewer FLOPs spent.
\begin{table*}
    \centering
    {\small
    \resizebox{\textwidth}{!}{
    \begin{tabular}{c|cc|cc|cc}
        \toprule
        \multirow{3}{*}{\textbf{Connector}} & \multicolumn{6}{c}{\textbf{$\Delta_{\texttt{Performance}} \ (\uparrow)$}}
        \\
        & \multicolumn{2}{c|}{\textbf{Multi-modal Image Classification}} & \multicolumn{2}{c|}{\textbf{Image-Text Matching}} & \multicolumn{2}{|c}{\textbf{Visual Question Answering}} \\
        & ImageNet-1K & CIFAR-100 & MSCOCO & Flickr-8K & OK-VQA & Text-VQA \\
        \midrule
        Linear & \textcolor{teal}{+}11.28 & \textcolor{teal}{+}10.62 & \textcolor{teal}{+}11.04 & \textcolor{teal}{+}11.14 & \textcolor{teal}{+}2.12 & \textcolor{teal}{+}2.29 \\
        MLP{$_1$} & \textcolor{teal}{+}4.50 & \textcolor{teal}{+}7.12 & \textcolor{teal}{+}2.08 & \textcolor{teal}{+}3.69 & \textcolor{teal}{+}0.24 & \textcolor{teal}{+}0.14 \\
        MLP{$_2$} & \textcolor{teal}{+}3.25 & \textcolor{teal}{+}6.21 & \textcolor{teal}{+}3.62 & \textcolor{teal}{+}4.55 & \textcolor{teal}{+}0.75 & \textcolor{teal}{+}0.24 \\
        \bottomrule
    \end{tabular}}}
\caption{\textbf{\textsc{Hyma} vs AutoPair Results ($N \times M=12$):} We show the performance gain ($\Delta$) of the best connector (for all connector configurations) obtained post stitching via \textsc{Hyma}, compared to that obtained via \textit{AutoPair}.}
\label{tab:sp}
\end{table*}
\paragraph{Image-text matching:} For image-text matching, we find higher values of Spearman's $\rho$, indicating that the stitches predicted by \textsc{Hyma} correlates strongly in performance with those obtained by full grid search on both MSCOCO and Flickr-8K. Similar to image-classification, we find that rank correlation metrics show more positive values for one dataset, Flickr-8K over the other, i.e., MSCOCO. In contrast, for image-text matching, we find that the performance gains (in recall$@5$) exhibited w.r.t \textit{Ask-LLM} baseline do not match those of \textit{UniT-1} in cases such as Linear connectors. In comparison to \textcolor{blue}{\textit{Oracle (GS)}}, average reduction in recall$@5$ is $3.03$ for MSCOCO and $2.52$ for Flickr-8K across all connectors.
\paragraph{Visual question answering:} In visual question answering on both OK-VQA and Text-VQA, Linear connectors exhibit the highest values in terms of NDCG $@ \ k$, Spearman's $\rho$, as well as performance gain. In line with the preceded evaluation tasks, i.e., multi-modal image classification and image-text matching, we find that connectors predicted by \textsc{Hyma} outperform those found by the \textit{Random, UniT-1} and \textit{Ask-LLM} baselines. Most notably, VQA emerges as the task with the least performance gap between \textsc{Hyma} and \textcolor{blue}{\textit{Oracle (GS)}}, with $0.88$ and $0.32$ being the difference in the respective recall$@$5 values across both datasets.
%
%
%
%
%
\subsection{\textsc{Hyma} vs AutoPair}
\label{app:AP}
We conduct a step-wise search-and-prune procedure over $6$ image encoders (evenly split across embedding dimensions $768$ and $1024$) and $2$ text encoders (also evenly split across embedding dimensions $768$ and $1024$). 
First we initialize a FLOPs budget equal to the total FLOP cost of searching over $N\times M=12$ pairs with \textsc{Hyma} for $10$ epochs. 
Next, our procedure trains connectors between all $12$ pairs for 2 epochs each, after which we rank each connector by its performance on a given task and dataset. After the ranking, we prune all pairs that exhibit performance that is less than or equal to the median performance. This is repeated until we exhaust the budget. If we are left with only one model after iterative pruning, we train it until the budget is exhausted. 

As shown in Table~\ref{tab:sp}, stitches obtained by AutoPair exhibit significantly lower performance than those obtained via \textsc{Hyma}, as the budget finishes before the individually trained connectors can reach strong performance. 
\subsection{Connection to Data Pruning}
\begin{table}[H]
\centering
    \resizebox{\linewidth}{!}{
    \begin{tabular}{@{}l c c@{}}
    \toprule
    \textbf{Method} & \textbf{Best Model configuration} & \textbf{Perf.} \\
    \midrule
    C-GS & DeiT-3S + miniLM-L & 24.07 \\
    \textsc{HYMA} & DeiT-3S + miniLM-L & \textbf{27.46} \\
    \bottomrule
\end{tabular}}
\caption{\textbf{\textsc{Hyma} vs. Constrained Grid Search (C-GS).} For the setting $N \times M = 3$, $B_m = 1$, we constrain the total data available to Grid Search to one-third, aligning it with \textsc{HYMA}’s data budget. While this constraint results in a comparable reduction in FLOPs relative to full Grid Search, it leads to a notable drop in performance. \textbf{Perf.} denotes MIC top-1 accuracy on ImageNet-1K.}
\label{tab:cgs}
\end{table}
While \textsc{Hyma} provides a unified and compute-efficient framework for addressing the M-OPS problem, the primary reduction in FLOPs arises from the dual mini-batching strategy employed during training. This dual mini-batching mechanism results in each model pair configuration being exposed to a smaller subset of data compared to independent stitching, effectively mimicking randomized data pruning in the process of constructing multimodal models from unimodal pairs.

Data pruning and filtering strategies for multimodal training have been extensively explored in prior work~\cite{fang2023datafilteringnetworks,bi2025prismselfpruningintrinsicselection,mahmoud2024sievemultimodaldatasetpruning}, typically focusing on restricting the training data via heuristic-based selection. In contrast, \textsc{Hyma} adopts a randomized approach: the mini-batching process dynamically selects data for each model configuration, and across multiple training steps, both the data and batch assignments are shuffled. This results in a more uniform and implicit allocation of the dataset across the space of possible model configurations, while still maintaining computational efficiency. It is important to note, however, that this data reduction applies only to each model configuration independently; the hypernetwork $H_\phi$, which generates the connector weights, is still trained over the entire dataset.

This effect is further evident when comparing \textsc{Hyma} to a constrained version of \textcolor{blue}{Oracle (Grid Search)} (C-GS). As shown in Table~\ref{tab:cgs}, when the total data available to Grid Search is limited to one-third—matching \textsc{Hyma}’s data budget—the best-performing model identified by C-GS performs significantly worse than \textsc{Hyma}. 
\section{Related Work}\label{sec:related_work}
\paragraph{Vision language models.} 
CLIP, one of the most popular VLMs, is contrastively pretrained on approximately 400M image-text pairs. Beyond multi-modal image classification and image-text retrieval, it has emerged to be applicable for tasks such as open-set  attribute recognition~\cite{ovarnet} and object detection~\cite{owlvit}. Moreover, it inspires modifications to the default InfoNCE recipe, such as image captioning with contrastive pretraining, using sigmoid in place of softmax on the InfoNCE similarity matrix, etc.~\cite{blip, flamingo, flava, siglip, conclip}. Additionally, datasets oriented towards CLIP-like vision-language pretraining have been released in recent times, including~\cite{laion, laion5b, yfcc100m, cc12m, desai2021redcaps}, often of the scale of millions of (image, caption) pairs. As a foundation model, CLIP has been applied in image synthesis~\cite{stablediff, dalle2}, and has been extended to modalities such as video~\cite{sdvideo, omnivl} and audio~\cite{audioclip}. Our work investigates how to efficiently develop multiple CLIP-like models from pretrained uni-modal encoder states.
%
\paragraph{Hypernetworks in LLMs and multi-modal domains.}
Hypernetworks~\citep{hypernetworks,schmidhuber1992learning} have been shown useful in improving training efficiency and adaptability in many machine learning pipelines~\cite{chauhan2024brief}.
Several works explored the advantages of hypernetworks for MLLMs and multi-modal models. Specifically, \cite{HyperLLaVA} proposes HyperLLaVA that predicts project parameters for MLLMs given  task input. Hypernetworks have also been used to predict the parameters of the adapters in parameter efficient fine-tuning of LLMs~\citep{mahabadi2021parameter,phang2023hypertuning} and VLMs \citep{Hyperpelt}. HyperCLIP~\citep{HyperCLIP}, trains a hypernetwork to predict the parameters of image encoder layers given the task. Overall, these models improve training efficiency and adaptability of a single combination on new tasks, but require grid search for more pairs. Our work addresses this limitation by training the joint hypernetwork for multiple encoders improving the efficiency and performance significantly.

\section{Conclusion}\label{sec:conclusion}
We present a novel investigation of the usage of hypernetworks for the M-OPS problem. \textbf{\textsc{Hyma}} is able to subvert expensive grid search across all uni-model model combinations, by learning connector parameters jointly, producing strongly initialised connectors. We demonstrate that \textsc{Hyma} is an efficient solution to the \textbf{M-OPS} problem. Also, \textsc{\textbf{Hyma}}'s design affords stitching of modalities beyond only image-text: other avenues include, for instance, audio-text. We hope to inspire future work that utilizes hypernetworks for similar problems, where training several small neural networks can be expressed as a generative model that learns the parameters of the target network. 
\section*{Limitations}
Hypernetwork training can be less stable than training a standard connector (i.e., a single MLP). Training instabilities in hypernetworks have been previously studied~\cite{ortiz2023magnitude,chauhan2024brief}, and are not unique to the specific design of our framework. However, since $H_\phi$ acts as a shared generating function across multiple connectors, the interaction of gradients from diverse model combinations—as well as their interplay with $B_m$—can still lead to instability during training. To stabilize training, we tune the $\beta_2$ parameter of the Adam optimizer in accordance with recommendations from the optimization literature~\cite{Catteneo_Adam}. In practice, we observed that including certain models (for example: the MaxVIT family~\cite{tu2022maxvit}) in the $N \times M$ pool led to instability, and thus these models were excluded from our final zoo. This limitation points to the need for a deeper investigation into the training dynamics and architectural properties of similar systems, which could inform strategies to improve both stability and performance of the hypernetwork.

\section*{Acknowledgements}
The authors thank the International Max Planck Research School for Intelligent Systems (IMPRS-IS) for supporting Diganta Misra. 
Jaisidh Singh is supported by the Konrad Zuse School of Excellence in Learning and Intelligent Systems (ELIZA) through the DAAD programme Konrad Zuse Schools of Excellence in Artificial Intelligence, sponsored by the Federal Ministry of Education and Research.
This work was enabled by compute resources provided by  Max Planck Institute for Intelligent Systems Tübingen \& Amazon Science Hub. 
\clearpage
\bibliography{custom}
%
%
\clearpage
\appendix
{\huge\noindent \hspace{75pt}\textbf{\textsc{Appendix}}}
\begin{figure*}[t]
    \centering
    \includegraphics[width=0.45\linewidth]{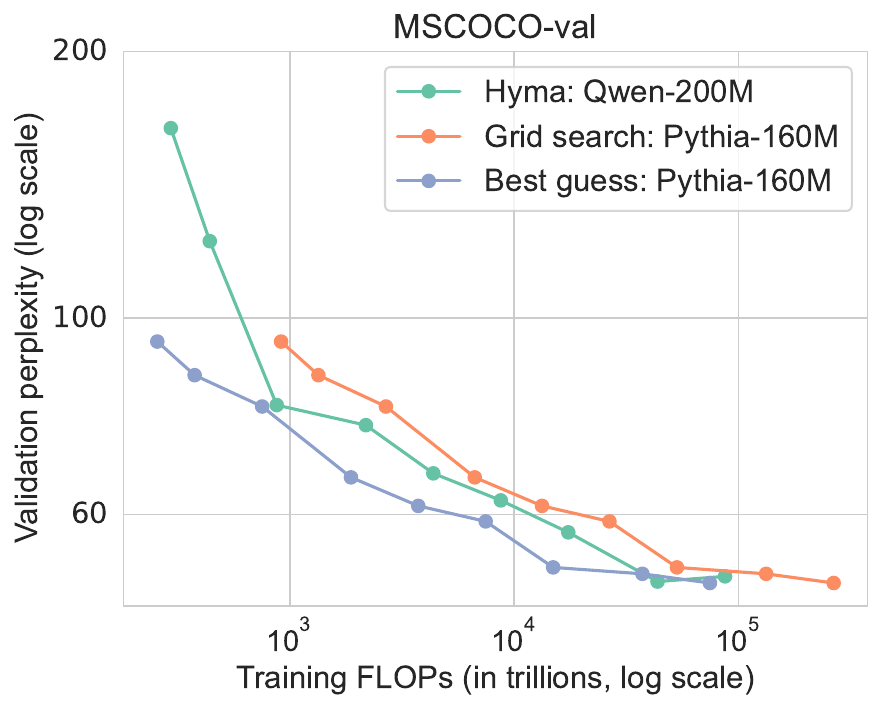}
    \includegraphics[width=0.45\linewidth]{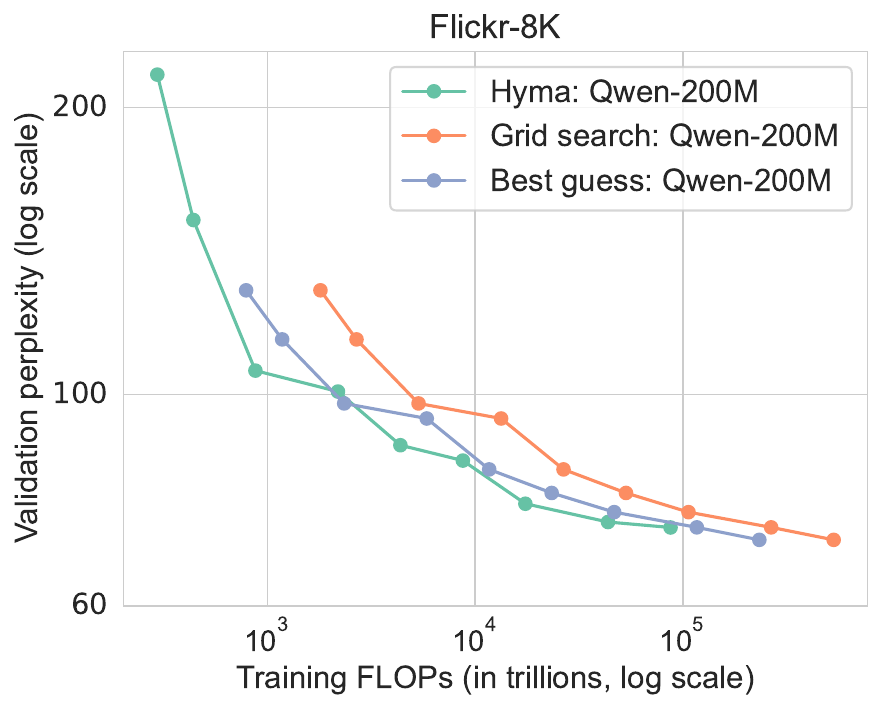}
    \caption{Evaluation of \textsc{Hyma} for MLLMs, on MSCOCO and Flickr-8K ($N=1,M=3,B_m=1$). We report the model combination exhibiting the best final performance for each evaluation benchmark and search method.}
    \label{fig:mllm_1_3}
\end{figure*}
\begin{table*}[!ht]
    \centering
    \small
    \resizebox{\textwidth}{!}{%
    \begin{tabular}{c|c|ccc|c|cccc}
    \toprule
        \multirow{2}{*}{\textbf{Dataset}} & \multirow{2}{*}{\textbf{Connector}} & \multicolumn{3}{c|}{\textbf{NDCG @ $k$ ($\uparrow$)}} & \textbf{$\rho$ ($\uparrow$)} & \multicolumn{4}{c}{\textbf{$\Delta_{\texttt{Perplexity}}$ ($\downarrow$)}} \\
       & & $k{=}5$ & $k{=}7$ & $k{=}9$ & $N{\times}M{=}9$ & Rand. ($n{=}5$) & UniT-1 & Ask-LLM & \textcolor{blue}{Oracle (GS)} \\
        \midrule
         \multirow{3}{*}{MSCOCO} & Linear & 0.16 & 0.42 & 0.74 & -0.6 & \textcolor{magenta}{+}3.68 & \textcolor{magenta}{+}6.85 & \textcolor{magenta}{+}3.20 & \textcolor{magenta}{+}6.85\\
         &  MLP$_1$ & 0.65 & 0.79 & 0.89 & 0.35 & \textcolor{magenta}{+}0.65 & \textcolor{magenta}{+}2.2 & \textcolor{magenta}{+}2.20 & \textcolor{magenta}{+}3.5 \\
         &  MLP$_2$ & 0.61 & 0.74 & 0.85 & 0.39 & \textcolor{magenta}{+}1.13 & \textcolor{magenta}{+}3.58 & \textcolor{magenta}{+}3.58 & \textcolor{magenta}{+}4.01 \\
        \midrule
         \multirow{3}{*}{Flickr-8K} & Linear & 0.56 & 0.73 & 0.85 & 0.12 & \textcolor{magenta}{+}1.65 & \textcolor{magenta}{+}5.54 & \textcolor{teal}{-}1.46 & \textcolor{magenta}{+}5.54 \\
         &  MLP$_1$ & 0.77 & 0.82 & 0.90 & 0.45 & \textcolor{teal}{-}0.1 & \textcolor{magenta}{+}1.30 & \textcolor{teal}{-}1.30 & \textcolor{magenta}{+}4.5 \\
         &  MLP$_2$ & 0.58 & 0.72 & 0.83 & 0.23 & \textcolor{teal}{-}3.57 & \textcolor{teal}{-}0.00 & \textcolor{teal}{-}0.00 & \textcolor{teal}{-}0.00 \\
        \bottomrule
    \end{tabular}%
    }
    \caption{\textbf{\textsc{Hyma} MLLM Results}: We report the ranking similarity between \textsc{Hyma} and the Oracle—Grid Search (GS)—using NDCG and Spearman’s $\rho$. Across all three connector configurations, \textsc{Hyma} exhibits strong correlation with GS rankings. Additionally, we show the perplexity difference ($\Delta$) of the best connector obtained post stitching via \textsc{Hyma}, compared to four baselines: (a) \textit{Random}: random pairing and stitching (avg. over 5 runs); (b) \textit{UniT-1}: stitching the best unimodal models; (c) \textit{Ask-LLM}: model pairs picked by Claude 4 Sonnet; and (d) \textit{Oracle}: Full grid search over all $N{\times}M{=}9$ configurations.}
    \label{tab:mllm_tab}
\end{table*}

\section{\textsc{Hyma} for Multi-modal Large Language Models (MLLMs)}\label{sec:mllm_exp}
Another avenue for employing a predictive model for stitching can be MLLMs, which is significantly different from the VLMs case. Not only is the causal language modeling objective different from the contrastive scheme of VLMs, the connector stitches output image encoder representations to LLM input representations. In VLMs, the connector strictly stitches output representations, i.e., features produced by the text encoder are stitched to the space of image encoder features. We are interested in investigating how \textsc{Hyma} responds to this setting via the following experiments.
\subsection{MLP$\mathbf{_1}$ | $\mathbf{N \times M=3}$} 
We stitch $N=1$ image encoder (\texttt{ViT-S}) with $M=3$ LLMs (\texttt{GPT-2~\cite{gpt2}, Pythia-160M~\cite{pythia}, Qwen-200M~\cite{qwen}}) using a 2-layer MLP (MLP$_1$) as the connector. Figure~\ref{fig:mllm_1_3} shows the performance of \textsc{Hyma} in comparison to the \emph{Grid Search} and \emph{Best Guess} baselines respectively. We report the performance of the best connectors identified by each search method, with the FLOPs incurred via training.
We find that \textsc{Hyma} reduces the cost of searching of all combinations, bringing it lower than training only one connector for the $N\times M=3$ case. The efficiency of \textsc{Hyma} over the two comparative baselines at the final state (rightmost point in each plot) is $3\times$ w.r.t. \emph{Grid Search} and $1.3\times$ w.r.t. \emph{Best Guess}. Further, all search methods yield comparable optimal perplexities, $51.1$ for \textsc{Hyma} and $51.0$ for \emph{Grid Search} (or \emph{Best Guess}) on MSCOCO. On Flickr-8K, the perplexities are found to be $72.4$ and $70.4$ for \textsc{Hyma} and \emph{Grid Search} (or \emph{Best Guess}) respectively. 
\subsection{Linear, MLP$_1$, MLP$_2$ | $\mathbf{N\times M = 9}$}
We scale up our experimental setting to now use $N=3$ image encoders (\texttt{Clip-ViT-B}, \texttt{DeiT3-B}, \texttt{ViT-S}) and $M=3$ LLMs (\texttt{GPT-2, Pythia-160M, Qwen-200M}). Similar to the case for VLMs, we vary the complexity of the connector from a linear layer, to an MLP with 2 hidden layers. Evaluation is done similarly to the case with $N\times M=3$ MLLM combinations, i.e., via image captioning on MSCOCO and Flickr-8K. As shown in Table~\ref{tab:mllm_tab}. \textsc{Hyma} struggles to match true ranking of model pairs for the MLLM case. Specifically, it performs worse on connectors of lower complexity, and consistently under-performs in terms of validation perplexity. Careful observation shows that for $N \times M = 9$ MLLMs, the ranking of connectors predicted by \textsc{Hyma} follows a trend of uni-modal model performance (the best image encoder (\texttt{Clip-ViT-B}) and LLM (\texttt{Qwen-200M}) show the best performance. However, independent stitching does not show such behavior. Overall, connectors obtained via independent stitching outperform those obtained from \textsc{Hyma} by a significant margin, and the true ranking diverges notably from that predicted by \textsc{Hyma}. Investigations on disentangling the effects of the causal modeling loss and the change in stitched representation spaces is left as future work.
\section{Pretrained models}
\label{sec:appendix}
\subsection{Image encoders (source: \texttt{timm}~\cite{timm})}
{\small
\begin{table}[H]
    \centering
    \resizebox{\columnwidth}{!}{
    \begin{tabular}{l|c|c|c|l}
    \toprule
        Feature & \multirow{2}{*}{Model} & \multirow{2}{*}{Shorthand} & Param. & \multirow{2}{*}{\texttt{timm} specifier}  \\
        Dim. & & & count (M) & \\
    \midrule
        \multirow{3}{*}{384} & \texttt{ViT-S} & \texttt{VS} & 22.05 & \texttt{vit\_small\_patch16\_224.augreg\_in21k\_ft\_in1k}\\
        & \texttt{DeiT-S} & \texttt{DS} & 22.05 & \texttt{deit\_small\_patch16\_224.fb\_in1k}\\
        & \texttt{DeiT-3S} & \texttt{D3S} & 22.06 &\texttt{deit3\_small\_patch16\_224.fb\_in1k}\\ 
        \midrule
        \multirow{3}{*}{768} & \texttt{ViT-B} & \texttt{VB} & 86.57 & \texttt{vit\_base\_patch16\_224.augreg\_in21k\_ft\_in1k}\\
        & \texttt{DeiT-B} & \texttt{DB} & 86.57 & \texttt{deit\_base\_patch16\_224.fb\_in1k}\\
        & \texttt{DeiT3-B} & \texttt{D3B} & 86.88 & \texttt{deit3\_base\_patch16\_224.fb\_in22k\_ft\_in1k}\\
        & \texttt{Clip-ViT-B} & \texttt{CVB} & 86.86 &  \texttt{vit\_base\_patch1\_clip\_224.laion2b\_ft\_in12k\_in1k}\\
        \midrule
        \multirow{3}{*}{1024} &\texttt{ViT-L} & \texttt{VL} & 304.33& \texttt{vit\_large\_patch16\_224.augreg\_in21k\_ft\_in1k}\\
        & \texttt{Eva2-L} & \texttt{E2L} & 305.08 & \texttt{eva02\_large\_patch14\_448.mim\_m38m\_ft\_in22k\_in1k}\\
        & \texttt{DeiT3-L} & \texttt{D3L} & 304.37 & \texttt{deit3\_large\_patch16\_224.fb\_in22k\_ft\_in1k}\\
    \bottomrule
    \end{tabular}}
    \caption{All pretrained image encoders used in our work are given above, along with their shorthand IDs that may be referred to in the main manuscript.\vspace{-6pt}}
    \label{tab:ie_tab}
\end{table}
}
\subsection{Text encoders \& LLMs (source: huggingface~\cite{huggingface})} 

{\small
\begin{table}[H]
    \centering
    \resizebox{\columnwidth}{!}{
    \begin{tabular}{l|c|c|c|l}
    \toprule
       Feature & \multirow{2}{*}{Model}  &  \multirow{2}{*}{Shorthand} & Param. &  \multirow{2}{*}{\texttt{huggingface} specifier} \\
       Dim. & & & count(M) & \\
       \midrule
        384 & \texttt{minilm-L} & \texttt{mlL} & 33.4 & \texttt{sentence-transformers/all-MiniLM-L12-v2}\\
        768 & \texttt{mpnet-B} & \texttt{mpB} & 109 & \texttt{sentence-transformer/all-mpnet-base-v2}\\
        1024 & \texttt{roberta-L} & \texttt{roL} & 355M & \texttt{sentence-transformer/all-roberta-large-v1}\\
        \midrule
        \multirow{3}{*}{768} & \texttt{GPT-2} & \texttt{g2} & 137 & \texttt{openai-community/gpt2}\\
        & \texttt{Pythia-160M} & \texttt{py} & 213 & \texttt{EleutherAI/pythia-160m}\\
        & \texttt{Qwen-200M} & \texttt{qw} & 203 & \texttt{MiniLLM/MiniPLM-Qwen-200M}\\
    \bottomrule
    \end{tabular}}
    \caption{All pretrained text encoders and LLMs used in our work are given above, along with their shorthand IDs that may be referred to in the main manuscript.\vspace{-6pt}}
    \label{tab:te_tab}
\end{table}
}
\lstset{style=mystyle}
\begin{figure*}[h!]
    \centering
    \begin{lstlisting}[language=Python, numbers=none]
    
    def train_hypernet(hypernet, data_iter, models_iter, optimizer, num_steps):
        hypernet.train()
        for step in range(num_steps):
            # first sample (image, caption) data with batch size B_d
            data_batch = next(data_iter)
            
            # then subsample the full NxM space of models with batch size B_m
            model_batch = next(models_iter)

            optimizer.zero_grad()

            # input to the hypernetwork are indices or ids of the respective pairs
            vlm_ids_in_full_zoo = get_ids_wrt_full_zoo(model_batch)
            
            # hypernet outputs parameters of the stitches between the pairs 
            generated_params = hypernet(vlm_pair_ids)
            
            # mapped the data through the stitched model pairs
            # and compute multi-pair multi-modal loss
            loss = hypernet.forward_data_through(
                data_batch,
                generated_params,
                model_batch
            )

            # back-propagate
            loss.backward()
            optimizer.step()
    \end{lstlisting}
    \caption{PyTorch~\cite{paszke2019pytorch} pseudocode for \textsc{HYMA} training procedure on $N \times M$ models.}
    \label{fig:pseudocode}
\end{figure*}
\section{Designing the Model Zoo}
While our empirical analysis suggests that models with larger parametric capacity or higher embedding dimensionality generally perform better after stitching, a natural question arises: why include smaller models in the model zoo at all? We justify their inclusion based on the following:
\begin{enumerate}
    \item First, including smaller models enables the construction of multi-modal models across a range of parametric capacities, which is crucial for deployment under varying computational or resource constraints. For example, an organization aiming to deploy multi-modal models at multiple scales would incur significantly higher training costs if relying on independent training for each configuration. In contrast, \textsc{HYMA} offers a substantially more cost-effective alternative.
    \item Second, our empirical observations indicate that larger models are not always the best-performing choice when stitched into multi-modal pairs. This motivates the inclusion of a diverse set of model configurations in our zoo to better explore the multi-modal design space. By covering a broader range of capacity combinations, \textsc{HYMA} facilitates a more comprehensive and efficient search, supported by observations from Figure~\ref{fig:motiv} and Table~\ref{tab:size_motivation}.
\end{enumerate}
\section{Training and hyper-parameter details}
\label{sec:hparams}
We tune hyperparameters for each trained model to maximize (i) validation performance, (ii) GPU utilization, and (iii) training stability. Our goal is to demonstrate that hypernetworks can efficiently approach the M-OPS problem that often requires a large amount of computational resources. Hence, we emphasize on the need to have maximum GPU utilization in order to present an efficiency-oriented method. We report the hyperparameters used for training connectors for VLMs along with the configuration for \textsc{\textbf{Hyma}}. We use $3$ random seeds and report average performance in each experiment. 
\paragraph{VLMs.} Training individual connectors between VLMs uses hyperparameters that provides the best performance after $10$ epochs of training. Our hyperparameter choice is similar to that of~\cite{ape}. Specifically, we use a batch size of $2^{14}$, the Adam optimizer, and a learning rate of $1e-2$ subject to a schedule that linearly warms up the learning rate from 0 for $50$ steps. After that, the learning rate is decayed to 0 following a cosine curve. Training \textsc{Hyma} for VLMs is quite sensitive to hyperparameters, as is to be expected from a complex network that outputs large spaces especially considering how it does so using indirectly (using layer-specific embeddings). The optimal batch size, i.e., that ensures the most stable training is $2^9$, and the learning rate is set to $1e-2$ for the Adam optimizer. As mentioned in the main manuscript, the value of the model batch size $B_m$ affect the training strongly, hence we set it to $1$ when $N\times M = 3$ and $9$ when $N\times M = 27$. For AutoPair, $N\times=12$ and $B_m=4$.
\paragraph{MLLMs.} For MLLMs, we follow recipes given in~\cite{tinyllava} for training only the connector (referred to as the feature alignment phase of pretraining). Particularly, we use Adam with batch size of $64$ for training individual connectors and learning rate $1e-3$. This is subject to a schedule of warmup ratio $3e-2$ following a cosine decay to 0. The batch size training \textsc{Hyma} for MLLMs is $32$ and the learning rate is $1e-3$.

\paragraph{Architectural experiments.} For VLMs, we tried using a compression of the image encoder features as the conditional input to the hypernetwork, while keeping all other components the same. Only the learnable code-book is replaced by a learnt compression of batch-averaged image encoder features. This configuration, denoted as \textsc{Hyma}$_{\text{EC}}$ yielded lower performance than our default methodology \textsc{Hyma}. Specifically for the $N\times M=3$ case, for multi-modal image classification on ImageNet-1K, we find that the top-1 accuracy of the best model pair given by \textsc{Hyma} is superior to that given by \textsc{Hyma}$_{\text{EC}}$ shown in Table~\ref{tab:ectab}.

\noindent We provide an example pseudo-code depicting our training setup in Figure~\ref{fig:pseudocode}.
\begin{table}[]
    \centering
    \begin{tabular}{c|c}
    \toprule
        Architecture & IN-1K Top-1 accuracy \\
        \midrule
        \textsc{Hyma} & \textbf{27.46}\\
        \textsc{Hyma}$_{\text{EC}}$ & 12.11\\
        \bottomrule
    \end{tabular}
    \caption{\textsc{Hyma} performs significantly better downstream in comparison to \textsc{Hyma}$_{\text{EC}}$}.
    \label{tab:ectab}
\end{table}
\section{Factors impacting FLOPs}
While the numbers of parameters in the model being trained is no doubt a factor that is linearly proportional to the total FLOPs incurred, we note that there are other factors like hyperparameters as well. For loss functions that relate linearly with the batch size, batch size has no effect on the total number of FLOPs incurred after the entire training run, as the model takes fewer update steps on a bigger batch size, but proportionately more on a smaller one. However, for loss functions that scale quadratically with the number of data samples observed, such as the InfoNCE loss~\cite{infonce}, the value of batch size can significantly affect the FLOP count. This, after the primary design choice of iteratively loading models, which decreases the number of samples shown to a model by $N\times M / B_m$, accounts for why \textsc{Hyma}, that training a large hypernetwork (of an average of $500\times$ more parameters than the connector) is efficient, particularly for VLMs. 
For the case of MLLMs, the reasons become our design choice of iterative model batches, as well as the fact that certain LLMs are of a larger parametric capacity than others. 
Hence backpropagating the gradient through them into the connector for a total of $\mathcal{T}$ steps is more expensive than doing so for $\mathcal{T} / (N \times M / B_m)$ steps via \textsc{Hyma}.
\section{Details of VQA implemention}
\label{app:vqa}
We follow a methodology similar to the method Question Irrelevant Prompt (QIP)~\cite{shen2021much,song2022clipmodelsfewshotlearners} that creates a prompt of ``QUESTION: \{\texttt{question}\} ANSWER: \{\texttt{answer}\}'' for a given image. This prompt is embedded via the text encoder and the task is to match the image to the correct prompt, as an image-text matching objective.

\section{Details of baselines}
\subsection{Ask-LLM (for VLMs)}
We prompt Claude 4 Sonnet~\cite{claude} to identify the best model pair:
\begin{tcolorbox}
    ``You are an oracle which will predict which combination of image and text encoders will perform best on a given task. The task is to predict which (image encoder, text encoder) pair will yield the best CLIP-like VLM from a list of image encoders and text encoders. More details about this: each pair of encoders will be connected via an MLP of number of hidden layers \{\texttt{depth}\} (0 means a linear layer), which will be trained to map text embeddings to the image embedding space such that the InfoNCE loss is minimized.\\ 

    \noindent Your job is NOT TO provide any code or run the experiment. JUST TO PREDICT WHICH PAIR WILL YIELD THE BEST \{\texttt{task}\} \{\texttt{task\_metric}\} on \{\texttt{dataset}\} (\{\texttt{dataset\_metadata}\}).\\

    \noindent Here are the image encoders, along with their metadata: \{\texttt{image\_encoders\_with\_metadata}\}\\

    \noindent Here are the text encoders, along with their metadata: \{\texttt{text\_encoders\_with\_metadata}\}\\

    \noindent Please provide your answer in (image\_encoder, text\_encoder) format ONLY. NO OTHER TEXT SHOULD BE PRODUCED BY YOU EXCEPT THE ANSWER IN THE REQUIRED FORMAT.''
\end{tcolorbox}
We specify the image encoder metadata from \texttt{timm} (details of the image encoder from the ImageNet-1K results database such as accuracy, parameters, image size for pretraining). The metadata of the text encoder is obtained via \texttt{huggingface} (details of the pretrained text encoder like embedding dimension, parameters). The ``\texttt{task}'' is one among multi-modal image classification, image-text matching, and visual question answering, whereas ``\texttt{dataset}'' is simply the name of the dataset, and ``\texttt{dataset\_metadata}'' contains the number of samples, classes, questions, and answers, as needed for the dataset. We specify the type of connector (Linear, MLP$_1$, MLP$_2$) via ``\texttt{depth}''. 
\subsection{Ask-LLM (for MLLMs)}
\begin{tcolorbox}
    ``You are an oracle which will predict which combination of image encoder and LLM will perform best on image captioning task. The task is to predict which (image encoder, LLM) pair will yield the best GPT4-like MLLM from a list of image encoders and LLMs. More details about this: each pair will be connected via an MLP of number of hidden layers \{\texttt{depth}\} (0 means a linear layer), which will be trained to map patch-wise image encoder outputs to the input embedding space of LLM such that the causal language modeling loss is minimized.\\ 

    \noindent Your job is NOT TO provide any code or run the experiment. JUST TO PREDICT WHICH PAIR WILL YIELD THE BEST \{\texttt{task}\} \{\texttt{task\_metric}\} on \{\texttt{dataset}\} (\{\texttt{dataset\_metadata}\}).\\

    \noindent Here are the image encoders, along with their metadata: \{\texttt{image\_encoders\_with\_metadata}\}\\

    \noindent Here are the LLMs, along with their metadata: \{\texttt{llms\_with\_metadata}\}\\

    \noindent Please provide your answer in (image\_encoder, llm) format ONLY. NO OTHER TEXT SHOULD BE PRODUCED BY YOU EXCEPT THE ANSWER IN THE REQUIRED FORMAT.''
\end{tcolorbox}
We specify image encoder details as done for VLMs, but LLMs details are obtained from \texttt{huggingface} (parameters, embedding dimension, context length).
%
%
%
\end{document}